%% file: iclr2026_conference.tex
\documentclass{article} 
\usepackage{iclr2026_conference,times}

\input{math_commands.tex}

\input{macros}

\usepackage{hyperref}
\usepackage{url}

\usepackage{graphicx}
\usepackage{algorithm}
\usepackage{algpseudocode}
\usepackage{multirow}
\usepackage{multicol}
\usepackage{array}
\usepackage{amsmath}

\usepackage{diagbox}
\usepackage{listings}
\lstdefinestyle{mystyle}{
    basicstyle=\footnotesize\ttfamily,
    backgroundcolor=\color{pink!20},  
    breaklines=true,                 
    breakatwhitespace=true,            
}
\usepackage{booktabs}
\usepackage{multirow}

\usepackage{caption}
\usepackage{xspace}

\usepackage[utf8]{inputenc}
\usepackage{pifont}
\usepackage{caption}
\usepackage{comment}
\usepackage{makecell}

\usepackage{amsthm}

\usepackage{tcolorbox}
\tcbuselibrary{breakable, skins}

\usepackage{wrapfig}
\usepackage{subcaption}  

\title{Dual-Scale World Models for LLM Agents towards Hard-Exploration Problems}

\author{Minsoo Kim \\
   Seoul National University \\
  \texttt{minsoo9574@snu.ac.kr} \\\And
  Seung-won Hwang\thanks{Corresponding author.} \\
  Seoul National University \\
  \texttt{seungwonh@snu.ac.kr} \\}

\iclrfinalcopy 
\begin{document}

\maketitle

\input{sections/abstract/abs_v1}
\input{sections/intro/intro_v1}
\input{sections/background/background_v1}
\input{sections/method/method_v1}
\input{sections/results/results_v1}
\input{sections/related/related_v1}

\input{sections/conclusion/conclusion_v1}

\input{reproducibility_statement}

\bibliography{iclr2026_conference}
\bibliographystyle{iclr2026_conference}

\appendix

\input{sections/appendix/mar_variance_reduction}

\section{Algorithms}
We provide the detailed overview of Go-Explore-based algorithms in Alg.~\ref{alg:ge_framework}, and the full algorithm of \ours in Alg.~\ref{alg:glow}.
\label{appendix:algorithms}
\input{algos/ours_mainalgo}
\input{algos/go_explore}

\input{sections/appendix/leakage}
\input{sections/appendix/api_cost}
\input{sections/appendix/prompts}

\section{Qualitative Examples}
\input{sections/examples/gwm_example}

\input{sections/examples/lwm_example}

\section{LLM Usage}
We utilized Claude for minor grammar and language edits in paper writing.

\end{document}

%% file: math_commands.tex
\usepackage{amsmath,amsfonts,bm}

\def\eqref#1{equation~\ref{#1}}

\def\1{\bm{1}}

\DeclareMathAlphabet{\mathsfit}{\encodingdefault}{\sfdefault}{m}{sl}
\SetMathAlphabet{\mathsfit}{bold}{\encodingdefault}{\sfdefault}{bx}{n}



%% file: macros.tex
\newcommand{\ours}{GLoW\xspace}
\newcommand{\oursfull}{\textbf{G}lobal-\textbf{Lo}cal \textbf{W}orld Models\xspace}

\newcommand{\ourslocalwm}{MAR\xspace}
\newcommand{\ourslocalwmfull}{Multi-path Advantage Reflection\xspace}

\definecolor{darkgreen}{rgb}{0.0, 0.5, 0.0}
\definecolor{extreme}{HTML}{B00000}
\definecolor{possible}{HTML}{009000}
\definecolor{difficult}{HTML}{C09000}

\newcommand{\xmark}{\ding{55}}%

\newcommand{\ci}[1]{{\scriptsize$\pm$#1}}

%% file: sections/abstract/abs_v1.tex
\begin{abstract}
LLM-based agents have seen promising advances, yet
they are still limited in ``hard-exploration'' tasks requiring 
\textit{learning new knowledge through exploration}. 
We present \ours, a novel approach leveraging dual-scale world models, maintaining a trajectory frontier of high-value discoveries at the global scale, while learning from local trial-and-error in exploration through a \ourslocalwmfull mechanism which infers advantage-based progress signals to guide exploration.
To evaluate our framework for hard-exploration, we tackle the Jericho benchmark suite of text-based games, 
where \ours achieves a new state-of-the-art performance for LLM-based approaches. 
Compared to state-of-the-art RL-based methods, our approach achieves comparable performance while requiring 100-800× fewer environment interactions.\footnote{Code will be open sourced at \url{https://github.com/mnskim/glow}}
\end{abstract}

%% file: sections/intro/intro_v1.tex
\section{Introduction}
While LLM agents~\citep{yao2023react,sumers2024cognitive,Wang_2024} excel at leveraging vast pre-trained knowledge in tasks such as robotic planning, software engineering, and web automation~\citep{ahn2022icanisay,yang2024sweagent,yang2025agentoccam},
they are reportedly limited in  \textit{hard-exploration problems}~\citep{Sutton1998, goexplore2019}. 
Hard exploration problems are typically characterized by large  state–action spaces, deceptive local optima, and sparse rewards. These factors often trap naive exploration in local optima, 
such that exploration fails to reach deeper states with rewards. For LLM agents, such problems pose two central challenges:
(1) Global learning, for maintaining long-term knowledge of valuable discoveries during exploration,
(2) Local trial-and-error, for quickly refining exploration policies from sparse environmental feedback.
Current LLM agent approaches
such as ReAct~\citep{yao2023react} or Reflexion~\citep{reflexion_shinn} support local trial-and-error, but lack mechanisms for long-term knowledge accumulation. Consequently, LLM agents fall short on hard-exploration tasks that humans can often solve effectively~\citep{cui2025tales,phan2025textquestsgoodllmstextbased}.

In this work, we introduce \oursfull (\ours), a framework for LLM agents that enables effective exploration in hard-exploration problems, by maintaining structured world models at 
two complementary scales for global and local learning.
Our approach builds on Go-Explore~\citep{goexplore2019} algorithm,
which achieves breakthroughs on hard-exploration problems by enhancing the exploration capabilities of RL and LLM-based agents~\citep{lu2025intelligent}.
The key idea of Go-Explore is to store discovered states into a \textit{state archive}:
Then, based on this archive, Go-Explore decomposes hard-exploration into alternating between:
(1) a \textit{selection} phase, choosing a \textit{promising} state from the archive to return to, and (2) an \textit{exploration} phase, to continue discovering new states from the selected state.
In its original implementation,  Go-Explore used hand-crafted heuristics for selection, and random action sampling for exploration, while later work, such as IGE~\citep{lu2025intelligent} improved selection to leverage LLM inference.
 
In this work, our core insight is that both selection and exploration require structured learning from past exploration experiences, but at different scales:
we first enrich beyond an
archive of isolated states, 
by additionally maintaining a trajectory frontier, which keeps the full temporal context of how high value states were reached and why progress stalled,
into a \textbf{global world model} for richer structured learning.
This allows an LLM-based analysis across the frontier to infer high-value regions as well as bottleneck states with high future potential,
enabling principled state selection in \ours, beyond heuristic or LLM-internalized notions of interestingness.
At the local scale, to guide exploration actions from the state,
we draw insights that advantage-based rewards better capture progress signals than Q-values~\citep{kazemnejad2025vineppo,setlur2025rewarding}: Our \ourslocalwmfull mechanism explores multiple trajectories from the same starting state and leverages LLM reasoning to infer \textit{intermediate advantages at key state-action pairs}. 
Through these advantage signals, the \textbf{local world model} enables controlled exploration under sparse environmental feedback.

To evaluate the capability of LLM agents in hard-exploration problems,
we study the Jericho benchmark suite of text-based games~\citep{hausknecht2019}, where SOTA has been
RL-based solutions~\citep{hausknecht2019,ammanabrolu2020graph,guo-etal-2020-interactive}
with $\varepsilon$-greedy or softmax exploration 
or MCTS-based exploration~\citep{jang2021montecarlo,shi2025monte}. 
However, they suffer from
poor sample efficiency, relying on extensive trial-and-error which requires \textbf{hundreds of thousands} of environment interactions.
Meanwhile, existing LLM agents were insufficient to address the challenge of learning from exploration
in Jericho games, showing limited performance compared to humans~\citep{cui2025tales,phan2025textquestsgoodllmstextbased}.

Through extensive experiments, 
we show that \ours 
improves the performance of LLM-based agents while achieving orders of magnitude improvement in sample efficiency compared to RL baselines. Our contributions are summarized as follows:
    \begin{itemize}
    \item We propose \ours, a novel LLM agent framework for hard-exploration problems through global-local world models,.
    \item We conduct comprehensive comparisons with existing agent approaches (RL, MCTS, LLM) and ablation studies to validate components of our method.
    \item We achieve a new state of the art for LLM-based approaches on Jericho, achieving comparable performance with RL-based SOTA, while reducing environment interactions required by 100-800×.
\end{itemize}    

%% file: sections/background/background_v1.tex
\section{Background}
\textbf{Jericho Benchmark} The Jericho benchmark (Hausknecht et al., 2019)
remains an unsolved hard-exploration problem,
where the text-based game environments provide two fundamental challenges~\citep{ammanabrolu2021modeling}: (1) partial observability, requiring agents to construct models of the world from local textual descriptions, and (2) combinatorial state-action spaces. 
For example in Zork1, the game vocabulary has 697 words and up to five-word commands, resulting in $O(697^5) = 1.64 \times 10^{14}$ possible actions per step, though only a tiny fraction are grammatically coherent and contextually relevant.
As a result, RL approaches, with simple exploration strategies, incur
hundreds of thousands interactions to offset sample inefficiencies in exploration.
This makes Jericho an ideal testbed for evaluating whether agents learn by exploring, rather than brute-force discovery.

\textbf{Methods for Hard-Exploration Problems}
Go-Explore~\citep{goexplore2019} achieved breakthroughs in hard-exploration problems by maintaining an archive of discovered states as global knowledge to (1) \textit{select} promising states
and (2) \textit{explore} from the state.
Algorithm~\ref{alg:ge_framework_skeleton} illustrates this high-level view, 
using example strategies from the original algorithm:
$selection$ returns the next state $s_{next}$, based on novelty driven heuristics (e.g., less visited states), and $explore$ generates actions 
(e.g., random action sampling in the original implementation), returning trajectory $\tau$. 
Appendix~\ref{appendix:algorithms} shows adaptations improve upon these heuristics. 
XTX~\citep{tuyls2022multistage} adapts imitation learning for selection and DQN for explore, and IGE uses LLM inference for both.
Beyond Go-Explore family, MCTS-based methods like MC-LAVE~\citep{jang2021montecarlo} and MC-DML~\citep{shi2025monte} leverage tree search with language-driven exploration and LLM priors respectively, though requiring 400,000+ interactions. 

\input{algos/go_skel}

%% file: algos/go_skel.tex
\begin{algorithm}[h]
\caption{Go-Explore Algorithm}
\label{alg:ge_framework_skeleton}
{\footnotesize
\begin{algorithmic}[1]
\Procedure{Go-Explore-Family}{$s_0$, $n_{iter}$}
    \State $\mathcal{A} \gets \{(s_0, \text{score}_0)\}$ \Comment{Archive}
    \For{$i = 1$ to $n_{iter}$}

    \State $s_{next} \sim select(\mathcal{A})  \propto \frac{1}{\text{visits}(s)^{\alpha}}$ \Comment{Novelty}
        \State   $\tau \gets explore(s_{next})  \propto \text{RandomActions}(s_{next})$ \Comment{No learning}
                \State Update $\mathcal{A}$
       \EndFor
\EndProcedure
\end{algorithmic}
}
\end{algorithm}

%% file: sections/method/method_v1.tex
\section{Method}
In this section, we describe the dual-scale learning paradigm of \ours in detail.

\subsection{Global World Model for State Selection}
The global world model extracts value signals from accumulated exploration trajectories. 
Unlike traditional state-based archives,
we maintain trajectories in a value-ranked frontier.
The global world model additionally maintains LLM-generated trajectory analysis.

\label{sec:method_selection}
\textbf{Value-Ranked Trajectory Frontier} As the source of value information, 
the global world model maintains a trajectory frontier $\mathcal{F} = \{\tau_1, \tau_2, ..., \tau_k\}$, containing the $k$ highest-value trajectories discovered during exploration, ranked by a value function $v : \mathcal{T} \rightarrow \mathbb{R}$. 
Each trajectory $\tau_i = (s_0^i, a_1^i, r_1^i, s_1^i, ..., a_T^i, r_T^i, s_T^i)$ represents a complete episode generated by the exploration policy $\pi_{\text{explore}}$ defined by the LLM agent, where $s_t \in \mathcal{S}$ are states, $a_t \in \mathcal{A}$ are actions, and $r_t \in \mathbb{R}$ are rewards. 
For the trajectory value function $v$, we use the maximum cumulative reward achieved during the episode, $v(\tau_i) = \max_{t \in [1,T]} \sum_{j=1}^{t} r_j^i$.
This is an effective choice for Jericho's sparse reward structure, and the possibility of encountering negative rewards or terminal failures.
In contrast to state-only representations, which lose the context of action and observation sequences, preserving complete trajectories enables accurate credit assignment and value estimation in sparse-reward environments where success depends on precise action sequences.

The frontier evolves progressively through iterative exploration. When exploration from selected states (detailed in Section~\ref{sec:method_exploration}) produces trajectory $\tau_{\text{new}}$ with value $v(\tau_{\text{new}})$, the frontier is updated:
\begin{equation}
\mathcal{F}_{t+1} = \text{top-}k(\mathcal{F}_t \cup \{\tau_{\text{new}}\}, v)
\end{equation}
This sliding window mechanism ensures the frontier maintains diverse high-value strategies, while allowing newly discovered superior trajectories to replace outdated ones. 
For any state $s_i$, we can derive the achieved value $v(s_i) = \max_{\tau \in \mathcal{F}, s_i \in \tau} v(\tau)$, representing the maximum value reached from state $s_i$ across all frontier trajectories. 
By tracking complete trajectories, the frontier serves as both an estimator of achieved values and a repository of successful action sequences.

\textbf{Motivation: Decomposing value for \textit{select} and \textit{explore}}
Inspired by UCB's value decomposition which balances exploitation with exploration bonus as:
\[
 \bar{V}(s) + c\sqrt{\frac{\log(N)}{n_s}}
\]
where $\bar{V}(s)$ is the empirical mean value and the second term is the exploration bonus based on visit count $n_s$,
we annotate two types of values $v$ and $v'$, corresponding to each term, by analyzing patterns across all frontier trajectories $\mathcal{F}$, to extract a set of critical global states with value annotations: 
\begin{equation}
W_{\text{global}} = g_{\text{LLM}}(\mathcal{F}) = 
\{(s_1, v_1, v'_1), (s_2, v_2, v'_2), \ldots, (s_k, v_k, v'_k)\}
\end{equation}
Here, each $(s_i, v_i, v_i')$ represents a key state identified from frontier analysis by a prompted LLM $g_{LLM}$, 
$v_i$ denotes the achieved value from $s_i$, 
while $v_i'$ reflects LLM's estimate of future value potential.
Importantly, this potential value $v_i'$ cannot be derived from trajectory scores alone, requiring LLM's reasoning about why trajectories fail and what progress could be achieved by resolving current bottlenecks.
For instance, a state where multiple trajectories fail might have \textit{low achieved value}, but have \textit{high potential value} when: (1) multiple high-value trajectories converge but fail to progress further, suggesting unexplored regions beyond, (2) partial solution patterns indicate missing components, or (3) environmental hints suggest valuable areas remain undiscovered.
This implements a \textit{semantic} form of optimism under uncertainty~\citep{auer_ucb, brafman_rmax} where UCB uses statistical bonuses while we derive optimistic values from LLM analysis of bottlenecks.
See Appendix~\ref{appendix:example_gwm_zork1} for a full example of $W_{global}$ generated for Zork1.

\textbf{Balancing Exploitation and Exploration in State Selection} We maintain a state archive $\mathcal{A} = \{(s_i, \text{score}(s_i))\}$ containing discovered states with their achieved scores. Given $W_{global}$, we select the next exploration state $s_{next}$ by balancing achieved and potential values via LLM:
\begin{figure}[h]
\centering
\begin{minipage}{0.45\textwidth}
\begin{algorithm}[H]
\small
\begin{algorithmic}[1]
\Require Frontier $\mathcal{F}$, State archive $\mathcal{A}$
\Ensure Selected state $s_{next}$
\State $W_{global} \gets g_{LLM}(\mathcal{F})$ where
\Statex  $W_{global} = \{(s_1, v_1, v'_1), \ldots, (s_k, v_k, v'_k)\}$
\Statex \hspace{1em} $v_i$: achieved, $v'_i$: potential
\For{each state $s \in \mathcal{A}$}
    \State $score[s] \gets align_{LLM}(s, W_{global})$
\EndFor
\State $s_{next} \gets \arg\max_{s \in \mathcal{A}} score[s]$
\State \Return $s_{next}$
\end{algorithmic}
\end{algorithm}
\end{minipage}
\hfill
\begin{minipage}{0.53\textwidth}
\centering
\includegraphics[width=\textwidth]{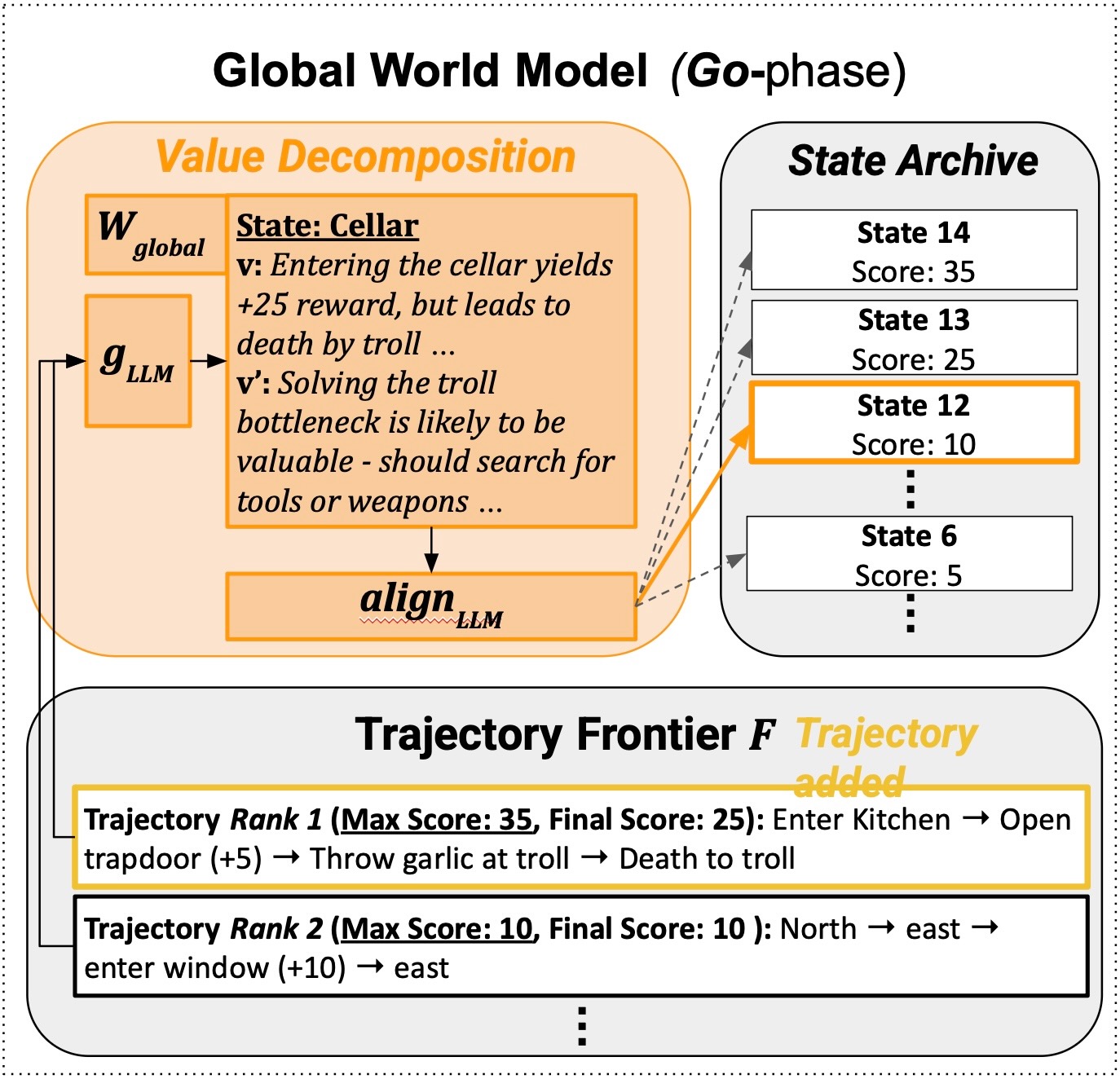}
\end{minipage}
\caption{\small(a) Select procedure in \ours, (b) Illustration of selection with Global World Model}
\label{fig:gwm}
\end{figure}
\vspace{-0.4cm}

where $\text{align}_{\text{LLM}}$ evaluates how well each archived state $s$ aligns with the high-value patterns identified in $W_{\text{global}}$. 
Since $W_{\text{global}}$ contains both achieved and potential values for key frontier states, this alignment naturally balances exploitation (favoring states similar to proven high-reward regions),
with exploration (prioritizing states near identified bottlenecks with high potential).
Fig.~\ref{fig:gwm} illustrates selection (\textit{l}.4 in Alg.~\ref{alg:ge_framework_skeleton}) in \ours with the Global World Model where a new trajectory (highlighted in gold) has been added to the frontier.
Once a state is chosen, we replay the stored sequence of actions to return to the state, which becomes the starting point of the next exploration phase, described in the following section.
 
\subsection{Local World Model for Exploration}
\label{sec:method_exploration}
In addition to the selection of states which align with exploration goals with high potential value, 
exploration can be enhanced by learning which actions are likely to lead to further progress, which is the objective of the local world model.

\textbf{Motivation: From Q-values to Advantages for Exploration} Existing LLM learning methods like self-reflection can be viewed as estimating state-action values (Q-values) from single trajectories. However, Q-value estimation from sparse rewards is notoriously high-variance~\citep{sutton2000, schulman2017proximalpolicyoptimizationalgorithms}, and we observe the same challenge in LLM-based learning: inferences from entire trajectories with sparse feedback are prone to incorrect causal attribution.

Drawing from RL theory, advantage functions $A(s,a) = Q(s,a) - V(s)$ reduce variance by comparing actions to a baseline rather than estimating absolute values. Recent work on process reward models (PRMs) further demonstrates that advantage-based rewards are more suited for exploration, by better capturing progress signals than Q-values, which tend to exploit known strategies~\citep{setlur2025rewarding, kazemnejad2025vineppo}. 

\textbf{Multi-path Advantage Reflection (MAR)} 
Inspired by
TRPO~\citep{pmlr-v37-schulman15}, computing robust advantage in sparse-reward setting over multiple rollouts from the same state,
we propose \ourslocalwmfull to compare
multiple trajectories from the same starting state, to produce pseudo-dense advantage signals from sparse environmental feedback. This effectively densifies the reward signal by inferring intermediate advantages at key state-action pairs, providing rich guidance for exploration where environmental rewards are insufficient.

Given a state $s$ selected by the global world model, we perform iterative exploration by sampling $n$ trajectories sequentially: after each trajectory $\tau_i$, we perform \ourslocalwm to extract learnings that inform the next trajectory $\tau_{i+1}$, in the form of world representation $W_{local}$. 
This creates a sequence $\{\tau_1, \tau_2, ..., \tau_n\}$ where each trajectory benefits from insights gained from previous attempts.
\begin{figure}[h]
\centering
\begin{minipage}{0.45\textwidth}
\begin{algorithm}[H]
\small
\begin{algorithmic}[1]
\Require Selected state $s_{next}$, Frontier $\mathcal{F}$, Exploration count $n$
\Ensure Trajectory set $\mathcal{T}_s$
\State $\mathcal{T}_s \gets \emptyset$
\State $W_{local} \gets \emptyset$
\For{$i = 1$ to $n$}
    \State $\tau_i \gets \pi_{explore}(s_{next}, W_{local}, \mathcal{T}_s, \mathcal{F})$
    \State $\mathcal{T}_s \gets \mathcal{T}_s \cup \{\tau_i\}$
    \State $W_{local} \gets \ourslocalwm(\mathcal{T}_s, \mathcal{F})$
    \Statex where $W_{local} = \{(s^*_1, A_{s^*_1}), ..., (s^*_k, A_{s^*_k})\}$
\EndFor
\State \Return $\mathcal{T}_s$
\end{algorithmic}
\end{algorithm}
\end{minipage}
\hfill
\begin{minipage}{0.53\textwidth}
\centering
\includegraphics[width=\textwidth]{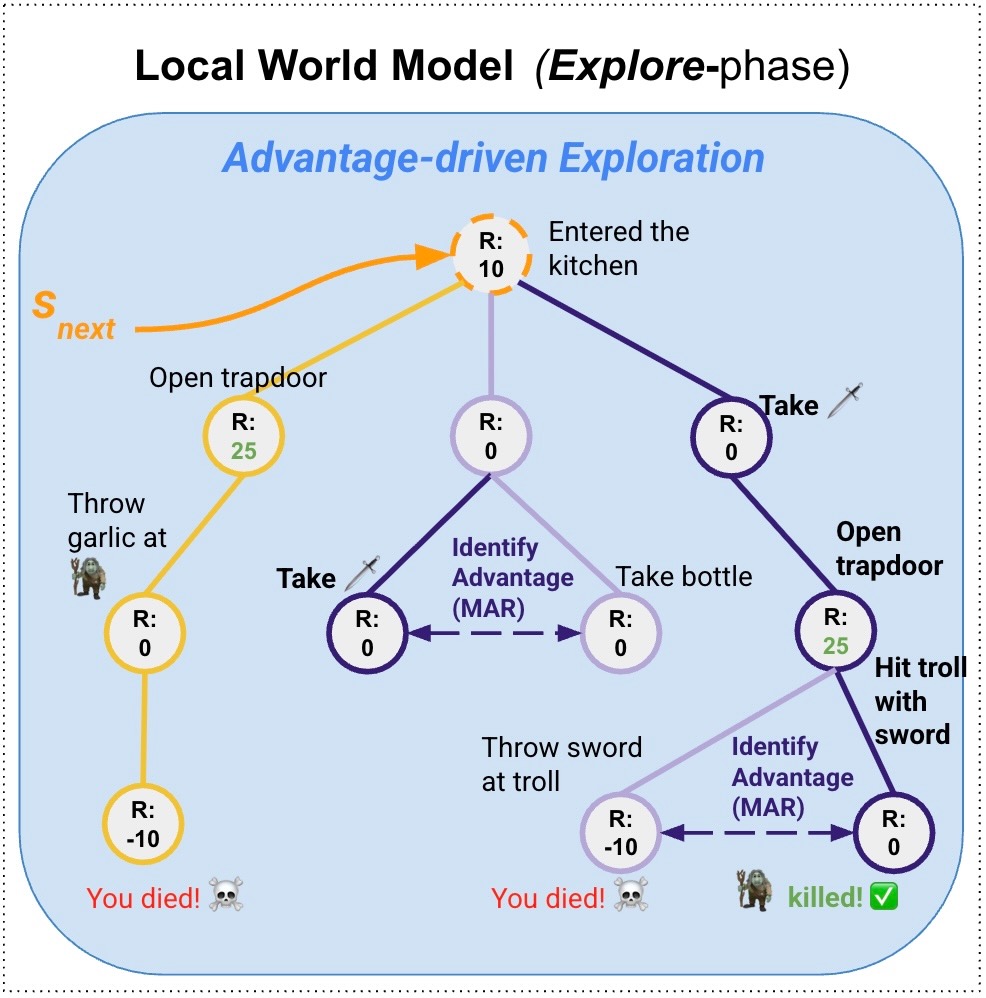}
\end{minipage}
\caption{\small (a) Explore procedure in \ours, (b) Illustration of exploration with Local World Model}
\label{fig:lwm}
\end{figure}
\vspace{-0.4cm}

where $\mathcal{F}$ provides global best trajectories as a stable value baseline, 
$\mathcal{T}_s = \{\tau_1, ..., \tau_n\}$ contains the trajectories sampled during the current exploration phase from state $s$.
States $s_1^*, ..., s_k^*$ are $k$ critical states (typically 2-4) 
where \ourslocalwm identifies valuable advantage information can be extracted,
either from divergent outcomes revealing good/bad actions, 
or from consistent patterns confirming reliable strategies.
\ourslocalwm focuses on these few decision points rather than annotating entire trajectories, enabling focused identification of which state-action pairs provide advantages.\footnote{In Appendix~\ref{sec:appendix_mar_theoretical}, 
we provide theoretical analysis showing how \ourslocalwm reduces variance through both multi-trajectory comparison and the stable baseline based on $\mathcal{F}$.
}

\textbf{Semantic Advantage Representation} Unlike scalar advantages $A(s,a)$,
\ourslocalwm produces $W_{\text{local}}$ containing rich \textit{semantic advantages} which encodes not just which actions are beneficial, but why they work and under what conditions, and captures progress signals which are not expressed by sparse rewards. See Appendix~\ref{appendix:example_lwm_zork1} for a full example of $W_{local}$ generated for Zork1.

\paragraph{Exploration Policy} The local world model enhances the \textit{explore} procedure in Alg.~\ref{alg:ge_framework_skeleton}($l$.5), by guiding a policy defined by an LLM agent, as: 
\label{sec:exploration_policy}
\begin{equation}
\pi_{\text{explore}}(a|s_t, h_t) = \text{Agent}_{\text{LLM}}(s_t, h_t, W_{\text{local}}, T_s, \mathcal{F})
\end{equation}
where $h_t$ is the current trajectory history, $T_s$ contains previous trajectories in the same exploration phase, and the policy leverages both learned advantages from $W_{\text{local}}$ and successful strategies from frontier $\mathcal{F}$.
Fig.~\ref{fig:lwm} illustrates exploration (\textit{l}.5 in Alg.~\ref{alg:ge_framework_skeleton}) in \ours with the Local World Model.
Consider a trajectory (gold) that reached the cellar but failed at the troll bottleneck without the sword. After analysis by the global world model (Fig.~\ref{fig:gwm}), which identifies high $v'$ at the cellar state, this state becomes $s_{next}$ (orange root, Fig.~\ref{fig:lwm}). The local world model drives multiple exploration attempts (purple paths), where \ourslocalwm identifies advantages for ``taking sword'' despite no immediate reward. This advantage learning guides successful exploration through the troll bottleneck (rightmost path).
To address Jericho's exponential action space, we implement a hybrid approach combining free generation with soft constraints.
While previous works use either constrained selection from valid actions in RL agents~\citep{hausknecht2019,ammanabrolu2020graph,tuyls2022multistage}
or pure free-form generation in LLM agents such as ReAct, we provide the valid actions to the LLM as a soft constraint, while still allowing free-form generation. This avoids failure modes of both approaches, where constrained selection can harm action diversity, while pure generation can produce many invalid actions.
As we show in Section 4.2, this hybrid approach, which we use consistently across both GLoW as well as all LLM baselines, significantly improves the base LLM performance with only a lightweight prompt and no few-shot examples.

%% file: sections/results/results_v1.tex
\section{Results}
We evaluate \ours on the Jericho benchmark suite, 
We present baselines (Sec.~\ref{sec:results_baselines}), setup (Sec.~\ref{sec:experimental_setup}),
main results demonstrating the effectiveness of \ours~(Sec.~\ref{sec:results_main}), and
ablation studies  (Sec.~\ref{sec:results_ablations}) isolating each module contribution.
Lastly, we provide detailed analysis of exploration dynamics in (Sec.~\ref{sec:analysis}).

\subsection{Baselines}
\label{sec:results_baselines}
We perform comprehensive comparison against baselines spanning RL-based, MCTS-based, and LLM-based approaches. 
Furthermore, we compare with specialized methods for hard-exploration problems in each type of baseline. 
All methods assume access to valid actions from Jericho.

\textbf{RL-Based Methods:}
\textbf{DRRN}~\citep{he-etal-2016-deep} is a value-based RL approach for choice-based games, learning Q-values for valid actions using GRU encoders and decoders trained via TD loss.
\textbf{KG-A2C}~\citep{ammanabrolu2020graph} is a on-policy RL agent that adapts Advantage Actor Critic (A2C)~\citep{mnih_a2c},
augmented by a dynamic knowledge graph as a state representation that is learned during exploration.
Similar to DRRN, \textbf{RC-DQN}~\citep{guo-etal-2020-interactive} is a DQN-based agent~\citep{mnih2015humanlevel}, but leverages object-centric neural reading comprehension architectures~\citep{seo2017bidirectional} for computing Q-values from observations.
\textbf{eXploit-Then-eXplore (XTX)}~\citep{tuyls2022multistage} is the current state-of-the-art method in Jericho, implementing Go-Explore with imitation learning on promising trajectories for state selection, and DQN with intrinsic curiosity reward for exploration. 
RL-based methods rely on million-scale interaction data to learn, 
leveraging parallel environments for training, with the exception of RC-DQN which leverages 100,000 interactions.

\textbf{MCTS-Based Methods:}
Monte Carlo Tree Search is widely adopted for large sequential decision-making problems~\citep{mcts_survey,alphago}, which explores effectively by combining random sampling and tree search.
\textbf{MC-LAVE}~\citep{jang2021montecarlo} combines MCTS with language-driven exploration, concentrating search effort on promising actions identified based on value estimates from semantically similar past actions.
\textbf{MC-DML}~\citep{shi2025monte} 
enhances MCTS by incorporating LLMs as action priors in the PUCT algorithm~\citep{alphago}, which balances exploration and exploitation during tree search. The LLM is equipped with a cross-trial memory mechanism, allowing it learn from past experiences such as death in Zork1.
Both methods require around 400,000 environment interactions to build comprehensive search trees.

\textbf{LLM-Based Methods:}
\textbf{ReAct}~\citep{yao2023react} is the widely adopted standard LLM agent approach interleaving reasoning and acting.
\textbf{Reflexion}~\citep{reflexion_shinn} is a multi-episode approach building on ReAct, incorporating self-reflection on each episode to guide future episodes.
\textbf{In-context Reinforcement Learning}~(ICRL)~\citep{song2025rewardenoughllmsincontext} is another multi-episode approach leveraging in-context reinforcement learning, using cumulative history of past trajectories and rewards as context for future episodes.
\textbf{Intelligent Go-Explore (IGE)}~\citep{lu2025intelligent} implements Go-Explore with LLMs, leveraging LLM-based state selection from a state archive, combined with ReAct-based exploration.
As LLM-based baseline methods  were not originally applied on Jericho, we re-implement them for Jericho using the action generation approach with valid action soft-constraint described in Sec.~\ref{sec:method_exploration}.
All LLM-based approaches use 1,000 interactions to balance performance and API cost.\footnote{We provide details of LLM API usage and cost in Appendix~\ref{sec:appendix_apicost}.}

\subsection{Experimental Setup} 
\label{sec:experimental_setup}
\textbf{Implementation Details} 
Each method is evaluated over 3 random seeds, reporting mean and standard deviation of maximum achieved scores. 
ReAct performs 20 independent 50-step episodes. Reflexion performs 20 trials of 50-step episodes, incorporating sliding-window memories from up to 10 previous attempts. Likewise, ICRL includes a sliding window of 10 previous trajectories as in-context examples. IGE and GLoW adaptively alternate between state selection and 50-step exploration episodes within the total 1,000 step budget. We use temperature 0.5 for all methods except IGE, which uses 0.3 following~\citet{lu2025intelligent}.
For \ours hyperparameters, $n$=3 exploration trajectories and $k$=5 trajectory frontier size is used. 

\textbf{Evaluation} We evaluate on 10 games from the Jericho benchmark~\citep{hausknecht2019}, spanning different difficulty levels. 
Following the benchmark's categorization, 
we test on \textit{possible games} (Pentari, Detective, Temple, Ztuu) featuring moderate puzzles and frequent rewards,
\textit{difficult games} (Zork1, Zork3, Deephome, Ludicorp) requiring more complex inventory management, puzzle-solving and navigation,
and \textit{extreme games} (Enchanter) involving non-standard actions and spell mechanics.
We use the standard Jericho interface providing textual observations and access to valid actions at each step.
Unlike some prior work, we do not augment observations with explicit ``look'' or ``inventory'' commands, instead allowing agents to learn these through play.

\input{tables/ours/ours_main_full}
\subsection{Main Results}
\label{sec:results_main}
We report our main results in Table~\ref{tab:ours_main}.
\ours achieves a new state-of-the-art performance among LLM approaches across 7 out of 10 games. 
On Zork1, a canonical game of the Jericho suite, our method reaches a score of 73.0,
a significant improvement over the next best LLM method (ICRL at 51.7), 
and surpassing all compared approaches (with the exception of XTX), including RL and MCTS baselines that use orders of magnitude more interactions. 
We observe the same strong improvements over the closest LLM method in Ludicorp (73.7 vs. 32.0 for ICRL), Enchanter (61.7 vs. 50.0 for IGE), Ztuu (29.3 vs. 18.7 for ReAct), and Balances (16.7 vs. 11.7 for ICRL).

Notably, our implementation of baselines with hybrid action generation approach shows  surprisingly strong performance, whereas prior works reported near-zero scores for LLM agents on Jericho~\citep{shi2025monte,cui2025tales,phan2025textquestsgoodllmstextbased}. 
Our implementation enables ReAct, Reflexion and ICRL to reach 48.3, 48.0, 51.7 on Zork1, respectively, and similarly on par with RL baselines such as KG-A2C and RC-DQN across the board. 
While this reveals the sample efficiency of LLM agents, 
these baselines still fall far short of more advanced exploration methods such as XTX and MC-DML, demonstrating the necessity of effective exploration for LLM agents.

Next we compare \ours against advanced exploration approaches. 
First, comparing with IGE which is the most directly comparable to ours as an LLM-based Go-Explore method, 
\ours substantially outperforms with better performance on 8 out of 10 games. 
\ours also achieves competitive performance with state-of-the-art RL and MCTS methods, XTX and MC-DML. 
We nearly match the overall state-of-the-art XTX, which uses 800× more interactions, on both Deephome (75.0 vs. 77.7) and Ludicorp (73.7 vs. 78.8), and notably surpass it on Enchanter (61.7 vs. 52.0). 
It also outperforms MC-DML, which employs extensive MCTS-based exploration around 400× more interactions, 
on most games including Zork1 (73.0 vs. 48.66), Deephome (75.0 vs. 67.0), and Ludicorp (73.7 vs. 19.67).
These results demonstrate that our dual-scale approach combining global world models for value-based state selection, with advantage learning for exploration, enables significant performance gains in LLM agents, competitive with sample-intensive RL approaches.

\subsection{Ablation Study}
\label{sec:results_ablations}
To validate the contribution of each component of \ours, we perform systematic ablations and report the results in Table~\ref{tab:ablation}.

\textbf{Effectiveness of Local World Model} 
We first analyze the efficacy of our local world model by ablating \ourslocalwm. We replace \ourslocalwm by Reflexion, which performs the same multi-path exploration but does not leverage our proposed advantage learning, 
instead performing single-trajectory reflection on the latest trajectory.
The results show that the performance drops significantly across most games, demonstrating that \ourslocalwm's advantage-based formulation more effectively leverages multi-trajectory information than Reflexion, improving exploration under sparse rewards.

\textbf{Effectiveness of Global World Model} 
Next, we analyze the effectiveness of the global world model, which consists of the frontier of high-value trajectories, and the LLM-based value analysis and alignment state selection.
We first ablate the LLM-based value analysis $W_{global}$, leveraging the raw frontier trajectories for state selection. 
The negative performance impact shows that, using LLM to reason across the frontier trajectories to infer potential value is indeed effective.
Next, we ablate the trajectory frontier $\mathcal{F}$ altogether, such that it is not used for state selection or leveraged by the exploration policy.
This causes further decrease in performance, confirming the contribution of the trajectory frontier in both phases.

\textbf{Synergy of LWM and GWM} Finally, we ablate all the above components together. The resultant model is similar to IGE, with multi-path Reflexion for exploration.
The results show that simply adding multi-path reflection does not lead to a clear improvement over IGE, indicating that the overall performance of \ours comes from the complementary synergy of its components.

\input{tables/ablation/ablation1}

\subsection{Analysis}
\label{sec:analysis}
\input{sections/analysis/vine}

%% file: tables/ours/ours_main_full.tex
\begin{table}[h!]
\centering
\resizebox{\textwidth}{!}{%
\renewcommand{\arraystretch}{1.2}
\begin{tabular}{@{}l|cccc|cc|cccc|c@{}}
\toprule
\multirow{2}{*}{Games} &
  \multicolumn{4}{c|}{\textit{RL-based}} &
  \multicolumn{2}{c|}{\textit{MCTS-based}} &
  \multicolumn{5}{c}{\textit{LLM-based}} \\ \cmidrule(l){2-12} 
 & \textbf{DRRN} & \textbf{KG-A2C} & \textbf{RC-DQN} & \textbf{XTX} & 
     \textbf{MC-LAVE} & \textbf{MC-DML} &
     \textbf{ReAct} & \textbf{Reflexion} & \textbf{ICRL} & \textbf{IGE} & \textbf{\ours (Ours)} \\ \midrule
Steps & 1,000,000 & 1,600,000 & 100,000 & 800,000 & $\sim$400,000 & $\sim$400,000 & 1000 & 1000 & 1000 & 1000 & 1000 \\ \midrule
\multirow{1}{*}{\textcolor{extreme}{Enchanter}}
  & 20 & 12.1 & 20 & \underline{52.0} & -- & 20\ci{0.0} & 46.7\ci{9.4} & 48.3\ci{9.4} & 43.3\ci{8.5} & 50.0\ci{7.1} & \textbf{61.7}\ci{20.1} \\ \midrule
\multirow{1}{*}{\textcolor{difficult}{Zork1}} 
  & 32.6 & 40.2\ci{0.4} & 38.8 & \textbf{103.4\ci{10.9}} & 45.2 & 48.66\ci{1.89} & 48.3\ci{4.7} & 48.0\ci{5.0} & 51.7\ci{4.7} & 44.3\ci{0.5} & \underline{73.0\ci{4.5}}  \\ \midrule
\multirow{1}{*}{\textcolor{difficult}{Zork3}}
  & 0.5 & 0.0 & 2.83 & \underline{4.2\ci{0.1}} & -- & 3\ci{0.0} & 3.0\ci{0.0} & 2.7\ci{0.5} & 3.0\ci{0.0} & 3.7\ci{0.9} & \textbf{4.3}\ci{0.9} \\
 \midrule
\multirow{1}{*}{\textcolor{difficult}{Deephome}}
  & 1 & 20\ci{2.1} & 1 & \textbf{77.7\ci{2.1}} & 35 & 67\ci{1.41} & 11.0\ci{4.2} & 22.0\ci{1.6} & 24.0\ci{5.7}  & 71.3\ci{4.9} & \underline{75.0\ci{8.7}} \\
 \midrule
\multirow{1}{*}{\textcolor{difficult}{Ludicorp}}
  & 13.8 & 19.8\ci{1.0} & 17 & \textbf{78.8} & 22.8 & 19.67\ci{1.7} & 19.7\ci{0.9} & 21.7\ci{1.2} & 32.0\ci{7.1} & 28.3\ci{11.3} & \underline{73.7\ci{11.0}} \\
  \midrule
\multirow{1}{*}{\textcolor{difficult}{Balances}}
  & 10 & 10 & 10 & \textbf{24} & 10 & 10\ci{0.0} & 10\ci{0.0} & 10\ci{0.0} & 11.7\ci{2.4} & 10.0\ci{0.0} & \underline{16.7\ci{2.4}} \\
  \midrule
\multirow{1}{*}{\textcolor{possible}{Pentari}}
  & 27.2 & 44\ci{0.9} & 43.8 & 49.6 & \underline{68} & \textbf{70\ci{0.0}} & 30.0\ci{0.0} & 30.0\ci{0.0} & 26.7\ci{4.7} & 30.0\ci{0.0} & 30.0\ci{0.0} \\ \midrule
\multirow{1}{*}{\textcolor{possible}{Detective}}
  & 197.8 & \underline{338\ci{3.4}} & 291.3 & 312.2 & 330 & \textbf{346.67\ci{9.43}} & 113.3\ci{4.7} & 166.7\ci{20.5} & 233.3\ci{47.8} & 316.7\ci{4.7} & 310.0\ci{8.2} \\ \midrule
\multirow{1}{*}{\textcolor{possible}{Temple}}
  & 7.4 & 8 & 8 & -- & 8\ci{0.0} & 8\ci{0.0} & 8.7\ci{0.9} & 8.7\ci{0.9} & 8\ci{0.0} & \textbf{13.7\ci{0.9}} & \underline{13.0\ci{0.0}} \\ \midrule
\multirow{1}{*}{\textcolor{possible}{Ztuu}}
  & 21.6 & 5\ci{0.0} & -- & -- & 7 & \underline{23.67\ci{1.9}} & 18.7\ci{2.4} & 18.3\ci{2.6} & 16.7\ci{4.1} & 15.0\ci{9.1} & \textbf{29.3\ci{4.0}} \\
  \bottomrule
\end{tabular}%
}
\caption{Comparison of RL-based, MCTS-based, and LLM-based methods on Jericho benchmark games. We report mean \ci{} standard error over 3 runs. \textbf{Bold} indicates best overall performance, and \underline{underline} indicates second-best. 
Steps shows total environment interactions. 
The color of game name indicates original game difficulty categories from \citet{hausknecht2019}: 
\textcolor{extreme}{extreme}, \textcolor{difficult}{difficult}, and \textcolor{possible}{possible}.
\ours achieves state-of-the-art among LLM-based approaches in 7/10 games, and is overall best among all compared approaches in 3/10, second-best in 5/10.
}
\label{tab:ours_main}
\end{table}
\vspace{-0.4cm}

%% file: tables/ablation/ablation1.tex
\begin{table}[h!]
\centering
\resizebox{1.0\textwidth}{!}{%
\renewcommand{\arraystretch}{1.2}
\begin{tabular}{@{}l|cccccc@{}}
\toprule
\textbf{Ablation Variants} & \textbf{Zork1} & \textbf{Zork3} & \textbf{Enchanter} & \textbf{Deephome} & \textbf{Ludicorp} & \textbf{Balances} \\
\midrule
\textbf{\ours (Full)} & \textbf{73.0\ci{4.5}} & \textbf{4.3\ci{0.9}} & \textbf{61.7\ci{20.1}} & \textbf{75.0\ci{8.7}} & \textbf{73.7\ci{11.0}} & \textbf{16.7\ci{2.4}} \\
\midrule
\xmark\xspace [Local WM] \ourslocalwmfull (\ourslocalwm) & 70.0\ci{13.6} & 4.3\ci{0.5} & 51.7\ci{9.4} & 56.7\ci{21.7} & 54.7\ci{22.4} & 11.7\ci{2.4} \\
\xmark\xspace [Global WM] State selection with $W_{global}$ & 62.0\ci{15.6} & 4.3\ci{0.9} & 60.0\ci{10.8} & 61.3\ci{26.0} & 63.3\ci{14.7} & 13.3\ci{2.4} \\
\xmark\xspace [Global WM] Trajectory frontier $\mathcal{F}$ & 61.7\ci{1.9} & 4.0\ci{0.8} & 53.3\ci{10.3} & 57.7\ci{23.3} & 63.3\ci{12.3} & 11.7\ci{2.4} \\
\xmark\xspace All above & 51.3\ci{5.2} & 4.3\ci{0.9} & 51.7\ci{9.4} & 56.0\ci{21.2} & 22.0\ci{0.8} & 10.0\ci{0.0} \\
\midrule
Standard IGE & 44.3\ci{0.5} & 3.7\ci{0.9} & 50.0\ci{7.1} & 71.3\ci{4.9} & 28.3\ci{11.3} & 10.0\ci{0.0} \\
\bottomrule
\end{tabular}%
}
\caption{Ablation study on \ours components. We evaluate the contribution of: (1) Local world model through \ourslocalwmfull, (2) Global world model for state selection, (3) trajectory frontier $\mathcal{F}$.
}
\label{tab:ablation}
\end{table}
\vspace{-0.4cm}

%% file: sections/analysis/vine.tex
\textbf{Controlling global vs local focus with $n$ exploration parameter}
We study the tradeoff between local learning depth and global exploration coverage by varying $n$, the number of explorations per selected state. Larger $n$ enables \ourslocalwm to learn from more trajectories, while smaller $n$ increases state selection frequency, helping escape local minima. With budget $B$=1000 and steps $s$=50, minimum state selections is $m = \lfloor B/(s \cdot n) \rfloor - 1$.
With $n$=1, \ourslocalwm is turned off. 
With $n$\textgreater1, \ourslocalwm analyzes $n$-1 local trajectories plus the global frontier trajectories.

\input{tables/ours/vine_scaling}

Table~(\ref{tab:vine_scaling}) shows that extreme values of $n$ generally yield suboptimal performance. 
When n=1, effectively disabling \ourslocalwm, performance drops significantly on certain games like Ludicorp (34.0 vs 73.7 with $n$=3). 
Conversely, Deephome shows consistent improvement with increasing $n$, suggesting it particularly benefits from deeper local exploration.
The results demonstrate that moderate increases in $n$ improve performance across several games, consistent with our theoretical analysis (Appendix~\ref{sec:appendix_mar_theoretical}) that \ourslocalwm benefits from variance reduction through multi-trajectory advantage calculation. 
However, setting $n$=5 begins to degrade performance, as excessive commitment to individual exploration phases reduces minimum state selection frequency to just 3, increasing susceptibility to local optima.
These findings indicate that balancing global and local learning is crucial.
We select $n$=3 as our default parameter, as it achieves the best overall performance by providing sufficient trajectories for robust advantage estimation while maintaining adequate state selection frequency to escape local minima.

%% file: tables/ours/vine_scaling.tex
\begin{table}[h]
\centering
\caption{Controlling the focus on global (less explorations per state but more frequent state selection) vs local learning (more explorations per state). The results demonstrate n=3 exploration from promising states strikes a good balance between the two.}
\label{tab:vine_scaling}
\resizebox{\textwidth}{!}{%
\renewcommand{\arraystretch}{1.2}
\begin{tabular}{lcccccccc}
\toprule
\multirow{4}{*}{\textbf{\makecell{Explorations\\per State}}} & 
\multirow{4}{*}{\textbf{\makecell{Max. \\Steps per\\Exploration\\Phase}}} & 
\multirow{4}{*}{\textbf{\makecell{Min. \\State\\Selection}}} & 
\multicolumn{6}{c}{\textbf{Score}} \\
\cmidrule(lr){4-9}
& & & Zork1 & Zork3 & Enchanter & Deephome & Ludicorp & Balances \\
& & & & & & & & \\[-0.5em]  
& & & & & & & & \\[0.5em]   
\midrule
1 (no \ourslocalwm) & $50\times1$ & 19 & 59.0\ci{5.7} & 3.7\ci{0.9} & 58.3\ci{9.4} & 59.7\ci{22.6} & 34.0\ci{15.6} & 13.3\ci{4.7} \\
2 (\ourslocalwm w/ 1) & $50\times2$ & 9 & 67.3\ci{8.7} & 3.7\ci{1.2} & 55.0\ci{7.1} & 43.3\ci{26.6} & 66.0\ci{3.7} & 11.7\ci{2.4} \\
\textbf{3} (\ourslocalwm w/ 2) & $50\times3$ & 5 & \textbf{73.0\ci{4.5}} & \textbf{4.3\ci{0.9}} & 61.7\ci{20.1} & 75.0\ci{8.7} & \textbf{73.7\ci{11.0}} & \textbf{16.7\ci{2.4}} \\
4 (\ourslocalwm w/ 3) & $50\times4$ & 4 & 63.0\ci{6.5} & \textbf{4.3\ci{0.9}} & \textbf{66.7\ci{10.3}} & 73.7\ci{4.5} & 62.0 \ci{12.4} & 16.7\ci{2.4} \\
5 (\ourslocalwm w/ 4) & $50\times5$ & 3 & 59.3\ci{13.8} & 4.0\ci{0.8} & 46.7\ci{6.2} & \textbf{76.3\ci{6.8}} & 53.3\ci{7.0} & 15.0\ci{0.0} \\
\bottomrule
\end{tabular}%
}
\vspace{-0.5em}
\end{table}

%% file: sections/related/related_v1.tex
\section{Related Works}

\textbf{Go-Explore-based Methods} Go-Explore~\citep{goexplore2019} enables effective exploration in sparse-reward environments by 
decomposing exploration into state selection and exploration
IGE~\citep{lu2025intelligent} adapts Go-Explore for LLMs, using LLM-based "promisingness" for state selection and ReAct for exploration. 
However, IGE's limited exploration and ill-defined selection criteria limit its effectiveness in complex environments like Jericho. Our work addresses these limitations through principled value decomposition for selection, and multi-path advantage learning for exploration.

\textbf{Agents for Text-based Games}
RL approaches to Jericho include DRRN~\citep{he-etal-2016-deep}, KG-A2C~\citep{ammanabrolu2020graph}, and RC-DQN~\cite{guo-etal-2020-interactive}, 
and the aforementioned XTX,
where all are sample-intensive, relying on hundreds of thousands of interactions. 
MCTS-based methods like MC-LAVE~\cite{jang2021montecarlo} and MC-DML~\cite{shi2025monte} leverage tree search but still rely on a similar scale of interactions.
We show that LLM agents can achieve comparable performance to RL methods,
while requiring orders of magnitude fewer interactions through structured exploration and learning mechanisms.

\textbf{Learning in LLM Agents} Recent works have studied how LLMs can learn from experience. Reflexion~\citep{reflexion_shinn} enables learning through self-reflection on failed attempts, while in-context reinforcement learning (ICRL)~\citep{song2025rewardenoughllmsincontext} leverages previous trajectories' history as context. However, these approaches struggle with sparse rewards due to noisy learning signals. Our \ourslocalwm mechanism addresses this challenge through multi-path advantage-based learning, providing more robust learning signals.

\textbf{World Models for LLM Agents} While traditional world models in model-based RL focused on transition dynamics~\citep{ha_world,hafner2024masteringdiversedomainsworld}, recent works show that in the context of LLMs, world models are usefully expanded as mechanisms for extracting task-sufficient state representations~\citep{tang2024worldcoder,li2024do}. 
Our dual-scale world models build on these insights, to learn both value patterns across global discoveries, and local advantage signals for exploration.

%% file: sections/conclusion/conclusion_v1.tex
\section{Conclusion}
We introduce \ours, a dual-scale world model framework to tackle hard-exploration problems.
\ours leverages a global world model that enables principled decomposition of state values,
and a local world model that integrates  trajectories from the same state as controlled exploration feedbacks.
Our approach achieves state-of-the-art performance among LLM methods on the challenging Jericho benchmark, while matching RL-based methods that require 800× more environment interactions.
By learning global value patterns across discoveries, and local progress signals from multi-path exploration, \ours overcomes a key limitation of LLM agents in hard-exploration tasks, demonstrating
a sample efficient yet high performance results.

%% file: reproducibility_statement.tex
\section*{Reproducibility Statement}
To ensure reproducibility of our results, we provide comprehensive implementation details in the paper. Algorithm~\ref{alg:glow} provides the complete pseudocode for GLoW, and hyperparameters are detailed in Section~\ref{sec:experimental_setup} ($n$=3 exploration trajectories, temperature=0.5, $k$=5 frontier size, 1000 environment steps). 
All prompts used for the global world model (Appendix~\ref{sec:appendix_prompts_gwm}),
LLM-based state selection (Appendix~\ref{sec:appendix_prompts_stateselection}),
MAR (Appendix~\ref{sec:appendix_prompts_lwm}), 
and exploration policy (Appendix~\ref{sec:appendix_prompts_actor}) are provided in full. 
Experiments use \texttt{GPT-4.1-mini-2025-04-14} as the LLM backbone, reporting results averaged over 3 random seeds with standard deviations. 
We implement all LLM baselines using the same action generation approach (Section~\ref{sec:exploration_policy}) for fair comparison. 
The Jericho benchmark is publicly available, and we use the standard evaluation protocol from Hausknecht et al. (2019). 
Code implementation will be publicly released upon publication.

%% file: sections/appendix/mar_variance_reduction.tex
\section{Theoretical Analysis of \ourslocalwmfull}
\label{sec:appendix_mar_theoretical}
\subsection{Variance Reduction in \ourslocalwm}

\textbf{Proposition 1.} \textit{Let \ourslocalwm analyze $n$ trajectories $\{\tau_1, ..., \tau_n\}$ starting from state $s$. For any critical state $s^*$ identified by \ourslocalwm, let $\hat{A}_{\text{single}}(s^*)$ denote an advantage estimate from analyzing a single trajectory through $s^*$, and $\hat{A}_{\text{\ourslocalwm}}(s^*)$ denote \ourslocalwm's advantage estimate from comparing $m \leq n$ trajectories that pass through $s^*$. Under the assumption of bounded variance across trajectories:}
\[
\text{Var}[\hat{A}_{\text{\ourslocalwm}}(s^*)] \leq \frac{\text{Var}[\hat{A}_{\text{single}}(s^*)]}{m}
\]

\textbf{Proof.} For a trajectory $j$ passing through state $s^*$ and taking action $a_j$, let $R_j(s^*, a_j)$ denote the random variable representing the sum of future rewards from $s^*$ onward. This provides an unbiased estimate of the true $Q(s^*, a_j)$.

The single-trajectory advantage estimate for action $a$ is:
\[
\hat{A}_{\text{single}}(s^*, a) = R_j(s^*, a) - \hat{V}(s^*)
\]
where $\hat{V}(s^*)$ is an estimate of the state value. This estimate has high variance because it relies on a single sample: $\text{Var}[\hat{A}_{\text{single}}(s^*, a)] = \text{Var}[R_j(s^*, a)]$ when $\hat{V}(s^*)$ is held constant.

Now consider MAR's approach. From the $m$ trajectories passing through $s^*$, let $m_a$ denote the number of trajectories taking action $a$. MAR computes an improved Q-value estimate by averaging outcomes:
\[
\hat{Q}_{\text{MAR}}(s^*, a) = \frac{1}{m_a} \sum_{j: a_j = a} R_j(s^*, a)
\]

Using basic properties of variance for independent random variables with equal variance $\sigma^2_a$:
\[
\text{Var}[\hat{Q}_{\text{MAR}}(s^*, a)] = \text{Var}\left[\frac{1}{m_a} \sum_{j: a_j = a} R_j\right] = \frac{1}{m_a^2} \cdot m_a \cdot \sigma^2_a = \frac{\sigma^2_a}{m_a}
\]

This shows variance reduction by factor $m_a$ for the Q-estimate. For the baseline in the advantage calculation, \ourslocalwm combines the global frontier trajectories $\mathcal{F}$, with local trajectories through $s^*$. The advantage estimate is:
\[
\hat{A}_{\text{MAR}}(s^*, a) = \hat{Q}_{\text{MAR}}(s^*, a) - \hat{V}_{\text{MAR}}(s^*)
\]

\ourslocalwm's hybrid baseline incorporating frontier trajectories serve a similar role to target networks in DQN~\citep{mnih2015humanlevel}. While target networks update parameters periodically to provide stable targets, our frontier baseline updates only when superior trajectories are discovered, providing stable value estimates that reduce learning instability.
Under the well-founded assumption that this stable baseline has low variance relative to the Q-estimate, the variance of the advantage estimate is dominated by the Q-component:
\[
\text{Var}[\hat{A}_{\text{MAR}}(s^*, a)] \approx \text{Var}[\hat{Q}_{\text{MAR}}(s^*, a)] = \frac{\sigma^2_a}{m_a} \leq \frac{\sigma^2_a}{1} = \text{Var}[\hat{A}_{\text{single}}(s^*, a)]
\]

More generally, for any action with $m_a \geq 1$ samples, we achieve variance reduction by a factor of $m_a$. This confirms that MAR reduces variance at each critical state, with greater reduction for actions sampled more frequently. $\square$

\textbf{Remark.} The proven variance reduction factor of $1/m$ represents a conservative lower bound for three reasons beyond the statistical averaging captured in the proof:

First, \ourslocalwm strategically identifies critical states $s_1^*, ..., s_k^*$ where advantage information is most valuable, rather than analyzing entire trajectories. This focused analysis avoids diluting the signal with irrelevant state transitions.

Second, the LLM provides semantic reasoning at these critical states, identifying causal patterns (e.g., lamp necessity for combat in darkness), generalizing across similar states, and leveraging prior knowledge, which are capabilities beyond pure statistical averaging.

Third, our sequential sampling with intermediate \ourslocalwm reflection means each $\tau_{i+1}$ benefits from analysis of $\{\tau_1, ..., \tau_i\}$, allowing later trajectories to avoid known failure modes and actively reduce uncertainty about critical decisions.

These enhancements explain why \ourslocalwm succeeds with small $m$ (typically 2-4 trajectories) in practice.

%% file: algos/ours_mainalgo.tex
\begin{algorithm}[h]
\caption{\ours: \oursfull}
\label{alg:glow}
{\footnotesize
\begin{algorithmic}[1]
\Procedure{\ours}{$s_0$, $n_{iter}$, $n_{explore}$, $k$}
    \State $\mathcal{F} \gets \emptyset$ \Comment{Initialize frontier}
    \State $\mathcal{A} \gets \{(s_0, 0)\}$ \Comment{Initialize state archive}
    \For{$i = 1$ to $n_{iter}$}
        \State $s_{\text{next}} \gets$ \Call{SelectState}{$\mathcal{F}$, $\mathcal{A}$}
        \State $\mathcal{T} \gets$ \Call{Explore}{$s_{\text{next}}$, $\mathcal{F}$, $n_{explore}$}
        \State \Call{UpdateArchive}{$\mathcal{T}$, $\mathcal{F}$, $\mathcal{A}$, $k$}
    \EndFor
    \State \Return $\arg\max_{\tau \in \mathcal{F}} v(\tau)$
\EndProcedure
\State
\Procedure{SelectState}{$\mathcal{F}$, $\mathcal{A}$}
    \State $W_{\text{global}} \gets g_{\text{LLM}}(\mathcal{F})$ 
    \State $s_{\text{next}} \gets \arg\max_{s \in \mathcal{A}} \text{align}_{\text{LLM}}(s, W_{\text{global}})$ \Comment{Select state based on decomposed value}
    \State \Return $s_{\text{next}}$
\EndProcedure
\State
\Procedure{Explore}{$s$, $\mathcal{F}$, $n$}
    \State $\mathcal{T} \gets \emptyset$ \Comment{Initialize trajectory set for current exploration phase}
    \State $W_{\text{local}} \gets \emptyset$ 
    \For{$j = 1$ to $n$}
        \State $\tau_j \gets \pi_{\text{explore}}(s, W_{\text{local}}, \mathcal{T}, \mathcal{F})$ \Comment{Rollout full trajectory from $s$}
        \State $\mathcal{T} \gets \mathcal{T} \cup \{\tau_j\}$
        \State $W_{\text{local}} \gets$ \Call{\ourslocalwm}{$\mathcal{T}$, $\mathcal{F}$}
    \EndFor
    \State \Return $\mathcal{T}$
\EndProcedure
\State
\Procedure{\ourslocalwm}{$\mathcal{T}$, $\mathcal{F}$}
    \State $W_{\text{local}} \gets f_{\text{LLM}}(\mathcal{T}, \mathcal{F})$ \Comment{Extract semantic advantages at key states}
    \State \Return $W_{\text{local}}$
\EndProcedure
\State
\Procedure{UpdateArchives}{$\mathcal{T}$, $\mathcal{F}$, $\mathcal{A}$, $k$}
    \For{$\tau \in \mathcal{T}$}
        \State $\mathcal{F} \gets \text{top-}k(\mathcal{F} \cup \{\tau\}, v)$ \Comment{Update the trajectory frontier}
        \For{$s' \in \tau$}
            \State $\mathcal{A} \gets \mathcal{A} \cup \{(s', score(s'))\}$ \Comment{Add states to state archive}
        \EndFor
    \EndFor
\EndProcedure
\end{algorithmic}
}
\end{algorithm}

%% file: algos/go_explore.tex
\begin{algorithm}[h]
\caption{Go-Explore-based Algorithms}
\label{alg:ge_framework}
{\footnotesize
\begin{algorithmic}[1]
\Procedure{Go-Explore-Family}{$s_0$, $n_{iter}$}
    \State $\mathcal{A} \gets \{(s_0, \text{score}_0)\}$ \Comment{Archive}
    \For{$i = 1$ to $n_{iter}$}
        \Statex \hspace{2em} \textit{--- Go Phase (State Selection) ---}
        \State \textbf{Go-Explore A:} $s_{next} \sim \text{Uniform}(\mathcal{A})$ \Comment{Random sampling}
        \State \textbf{Go-Explore B:} $s_{next} \sim P(s) \propto \frac{1}{\text{visits}(s)^{\alpha}}$ \Comment{Novelty}
        \State \textbf{Go-Explore C:} $s_{next} \sim P(s) \propto \text{domain}(s)$ \Comment{Domain heuristics}
        \State \textcolor{purple}{\textbf{XTX:} $s_{next} \gets \text{ImitationLearning}(\mathcal{T})$ \Comment{Imitation learning}}
        \State \textcolor{orange}{\textbf{IGE:} $s_{next} \gets \text{LLM.SelectPromising}(\mathcal{A})$ \Comment{Ill-defined promising-ness}}
        \State \textcolor{blue}{\textbf{GLoW:} $s_{next} = \text{align}_{\text{LLM}}(s, W_{\text{global}}) 
        $}
\Statex \textcolor{blue}{\hspace{3em} \Comment{Principled value decomposition (Sec.~\ref{sec:method_selection})}}
        \State 
        \Statex \hspace{2em} \textit{--- Explore Phase ---}
        \State \textbf{Go-Explore:} $\tau \gets \text{RandomActions}(s_{next})$ \Comment{No learning}
        \State \textcolor{purple}{\textbf{XTX:} $\tau \gets \text{DQN}(s_{next})$ \Comment{DQN with curiosity reward}}
        \State \textcolor{orange}{\textbf{IGE:} $\tau \gets \text{ReAct}(s_{next})$ \Comment{Standard LLM agent}} 
        \State \textcolor{blue}{\textbf{GLoW:} For $j = 1$ to $n$: \Comment{LLM agent with advantage-driven exploration} (Sec.~\ref{sec:method_exploration})} 
        \State \textcolor{blue}{\hspace{1em} $\tau_j \gets \pi(s_{next}, W_{\text{local}})$}
        \State \textcolor{blue}{\hspace{1em} $W_{\text{local}} \gets \text{\ourslocalwm}(W_{\text{local}}, \tau_j, \mathcal{F})$ } 
        \State 
        \Statex \hspace{2em} \textit{--- Archive Update ---}
        \For{each state $s'$ in $\tau$}
            \If{$\text{IsNotRedundant}(s', \mathcal{A})$} \Comment{Domain-specific novelty}
                \State $\mathcal{A} \gets \mathcal{A} \cup \{s'\}$
            \EndIf
        \EndFor
    \EndFor
\EndProcedure
\end{algorithmic}
}
\end{algorithm}

%% file: sections/appendix/leakage.tex
\section{Contamination Check}
\input{tables/contamination/contamination_gpt4.1_300walkthrough_update}
To assess whether large language models have prior knowledge of Jericho games, we conducted a data contamination analysis following the methodology of \citet{tsai2025largelanguagemodelsplay}.
We evaluate contamination by testing whether models can navigate between locations without being shown any gameplay.
Specifically, we: (1) collect a walkthrough trajectory by executing up to 300 steps from each game's built-in Jericho walkthrough actions,
(2) build a graph of locations and transitions from this walkthrough,
(3) generate navigation questions asking for paths between observed locations, and
(4) query the model with these questions without providing any context.
Navigation questions take the form: ``In [GAME], what steps would you take to go to [LOCATION B] from [LOCATION A]?''
We evaluate responses using strict pattern matching with word boundaries, requiring the exact sequence of navigation commands to appear consecutively in the model's response.

Table~\ref{tab:contamination_gpt4.1mini} shows results of contamination checks for \texttt{GPT-4.1-mini} across 11 Jericho games.
We observe minimal contamination, with all games showing below 20\% accuracy.
Most games (8 out of 11) show less than 10\% accuracy, consistent with random guessing or generic text adventure knowledge.
The slightly higher accuracies for Ludicorp (19.7\%), Deephome (17.0\%), and Library (15.4\%) likely reflect the model providing common navigation commands (e.g., "go south") that occasionally match by chance.
Even famous games like Zork1 (10.9\%) show accuracy near chance level, while less-known games like Balances (1.9\%) and Pentari (1.4\%) show essentially no prior knowledge.
These low accuracy rates, combined with the model's generic responses that lack game-specific details, indicate that our experimental results reflect genuine exploration and reasoning capabilities rather than memorized solutions.

%% file: tables/contamination/contamination_gpt4.1_300walkthrough_update.tex
\begin{table}[h]
\centering
\caption{Data contamination analysis: LLM accuracy (\%) on navigation questions without seeing gameplay.}
\label{tab:contamination_gpt4.1mini}
\resizebox{0.4\textwidth}{!}{%
\renewcommand{\arraystretch}{1.2}
\begin{tabular}{lcc}
\toprule
\textbf{Game} & \textbf{\# Questions} & \textbf{Accuracy (\%)} \\
\midrule
Zork1 & 230 & 10.9 \\
Zork3 & 194 & 8.2 \\
Enchanter & 239 & 9.2 \\
Detective & 66 & 9.1 \\
Balances & 54 & 1.9 \\
Library & 26 & 15.4 \\
Pentari & 70 & 1.4 \\
Deephome & 288 & 17.0 \\
Temple & 92 & 12.0 \\
Ludicorp & 320 & 19.7 \\
Ztuu & 71 & 9.9 \\
\bottomrule
\end{tabular}%
}
\end{table}

%% file: sections/appendix/api_cost.tex
\subsection{LLM API Cost}
\label{sec:appendix_apicost}
We use \texttt{gpt-4.1-mini-2025-04-14} for all LLM components (\$0.40/\$1.60 per million input/output tokens). Per-run costs of all LLM-based approaches with 1,000 environment steps range from \$4 to \$6, with negligible differences across approaches,
maintaining practicality for research iteration.

%% file: sections/appendix/prompts.tex
\section{Prompts}
We present the full prompts used in \ours. 
Our prompts rely solely on simple instructions and structured output formats without requiring few-shot exemplars, enabling the method to generalize across diverse game scenarios.

\subsection{Frontier Trajectory Analysis}
\label{sec:appendix_prompts_gwm}
\input{sections/prompts/gwm}

\subsection{State Selection}
\label{sec:appendix_prompts_stateselection}
\input{sections/prompts/state_selection}

\subsection{\ourslocalwmfull (\ourslocalwm)}
\label{sec:appendix_prompts_lwm}
The \ourslocalwm prompt generates $W_{\text{local}}$ as described in Section~\ref{sec:method_exploration}, identifying critical decision points and their associated advantages from multiple exploration trajectories. The prompt incorporates three inputs: (1) the global trajectory frontier containing highest-value trajectories that serve as value baselines, (2) local exploration attempts from the current phase showing different outcomes from the same starting state, and (3) previous $W_{\text{local}}$ outputs when available, enabling cumulative learning within the exploration phase.

By comparing outcomes across these trajectory sources, \ourslocalwm produces $W_{\text{local}} = \{(s^*_i, A_{s^*_i})\}^k_{i=1}$, identifying where specific actions provide clear advantages. This semantic representation captures causal relationships (e.g., ``taking the lamp enables combat in darkness'') rather than strictly scalar values, enabling the exploration policy to leverage both statistical patterns from trajectory comparison and LLM reasoning about game mechanics at critical states.

\input{sections/prompts/mar}

\subsection{Exploration Policy in \ours}
The prompt for the exploration policy in \ours is shown below.
We instruct the LLM to use a JSON output format with reasoning and action fields.
\label{sec:appendix_prompts_actor}
\input{sections/prompts/actor}

%% file: sections/prompts/gwm.tex
\begin{tcolorbox}[
    title={Analysis $(g_{\text{LLM}})$ Prompt},
    colback=gray!5!white,
    colframe=gray!75!black,
    fonttitle=\bfseries,
    breakable,
    enhanced,
    fontlower=\small\ttfamily,
    before upper=\small\ttfamily
]
Analyze these successful game trajectories to identify patterns and strategy:\\
\\
\textcolor{blue}{\{For each trajectory in $\mathcal{F}$:\}}\\
Trajectory N (Peak: X, Final: Y):\\
\hspace*{1em}[score] action -> observation (reward: +/-N if non-zero)\\
\hspace*{1em}[score] action -> observation\\
\hspace*{1em}...\\
\\
Based on these trajectories, provide a strategic analysis:\\
\\
1. FRONTIER \& EXPLORATION STATUS:\\
\hspace*{1em}- What areas/locations have been successfully reached?\\
\hspace*{1em}- What remains unexplored or inaccessible?\\
\\
2. GAME CHECKPOINTS \& PROGRESS:\\
\hspace*{1em}- What are the key milestones/checkpoints identified?\\
\hspace*{1em}- What items or abilities unlock new areas?\\
\hspace*{1em}- What phase of the game are we in?\\
\\
3. BOTTLENECKS \& CHALLENGES:\\
\hspace*{1em}- Where do trajectories commonly get stuck?\\
\hspace*{1em}- What obstacles block further progress?\\
\hspace*{1em}- What resources or knowledge are we missing?\\
\\
4. REWARD STRUCTURE:\\
\hspace*{1em}- When and how are points earned?\\
\hspace*{1em}- What actions yield the highest rewards?\\
\hspace*{1em}- Are there patterns to the scoring?\\
\\
5. NEXT INVESTIGATION GOALS:\\
\hspace*{1em}- What specific objectives should we pursue?\\
\hspace*{1em}- Which unexplored areas are most promising?\\
\hspace*{1em}- What items or states do we need to reach?\\
\\
Provide a concise strategic summary focusing on actionable insights.
\end{tcolorbox}

%% file: sections/prompts/state_selection.tex
\begin{tcolorbox}[
    title={State Selection $(\text{align}_{\text{LLM}})$ Prompt},
    colback=gray!5!white,
    colframe=gray!75!black,
    fonttitle=\bfseries,
    breakable,
    enhanced,
    fontlower=\small\ttfamily,
    before upper=\small\ttfamily
]
=== STRATEGIC GAME ANALYSIS ===\\
\textcolor{blue}{\{Analysis of frontier trajectories $W_{\text{global}}$\}}\\
==================================================\\
\\
Based on the above analysis, select the state from the archive that:\\
- Best aligns with the identified investigation goals\\
- Can help overcome identified bottlenecks\\
- Explores promising frontiers\\
- Has potential for high rewards based on patterns\\
\\
Current state archive:\\
\\
0: [Score: X, Steps: Y, Visits: Z]\\
\hspace*{1em}Observation: \textcolor{blue}{\{state observation\}}\\
\hspace*{1em}Inventory: \textcolor{blue}{\{state inventory\}}\\
\\
1: [Score: X, Steps: Y, Visits: Z]\\
\hspace*{1em}Observation: \textcolor{blue}{\{state observation\}}\\
\hspace*{1em}Inventory: \textcolor{blue}{\{state inventory\}}\\
\\
...\\
\\
Choose state index (0-N).\\
Respond in JSON format:\\
\{\\
\hspace*{1em}"thought": "Your reasoning about which state best aligns with the strategic goals",\\
\hspace*{1em}"index": <number>\\
\}
\end{tcolorbox}

%% file: sections/prompts/mar.tex
\begin{tcolorbox}[
    title={$W_{local}$ Generation Prompt (\ourslocalwm)},
    colback=gray!5!white,
    colframe=gray!75!black,
    fonttitle=\bfseries,
    breakable,
    enhanced,
    fontlower=\small\ttfamily,
    before upper=\small\ttfamily
]
Review these exploration attempts and identify KEY STATE ADVANTAGES:\\

\textcolor{blue}{\{Previous $W_{local}$ from earlier iterations, if any\}}\\

\textcolor{blue}{\{Global frontier trajectories $\mathcal{F}$\}}\\

\textcolor{blue}{\{Local exploration trajectories from state $s$\}}\\

Analyze all trajectories and identify ADVANTAGES at KEY STATES:\\

For each important location/state observed across ALL attempts, list:\\
- STATE: [description of state/location]\\
- ADVANTAGES discovered:\\
\hspace*{1em}• [specific action] → [specific benefit/outcome] (score impact if clear)\\
\hspace*{1em}• [what to avoid] → [consequence] (score impact if clear)\\
\hspace*{1em}• [optimal sequence] → [why it's better]\\

Example format:\\
STATE: At the house entrance with lamp\\
- ADVANTAGES:\\
\hspace*{1em}• "go east" → finds sword (enabled +15 points later)\\
\hspace*{1em}• "open mailbox first" → gets crucial map (+5 immediate)\\
\hspace*{1em}• avoid "go upstairs" early → wastes moves in empty attic (-7 overall)\\

Focus on:\\
1. States that appear across multiple attempts (to see different outcomes)\\
2. Critical decision points where scores diverged significantly\\
3. Action sequences that consistently led to success or failure\\
4. Items or information that enabled later progress\\

Provide 2-4 KEY STATES with their discovered advantages.\\
Be specific about actions, items, and locations from the actual game.
\end{tcolorbox}

%% file: sections/prompts/actor.tex
\begin{tcolorbox}[
    title={System Prompt},
    colback=gray!5!white,
    colframe=gray!75!black,
    fonttitle=\bfseries,
    breakable,
    enhanced,
    fontlower=\small\ttfamily,
    before upper=\small\ttfamily
]
You are exploring a text adventure game. Your goal is to make progress and increase your score.\\
\\
Generate actions that explore new possibilities and make progress.\\
\\
Respond in JSON format:\\
\{\\
\hspace*{1em}"thought": "Your reasoning about what to try",\\
\hspace*{1em}"action": "the exact command to execute"\\
\}
\end{tcolorbox}

\begin{tcolorbox}[
    title={User Prompt (at initial step)},
    colback=gray!5!white,
    colframe=gray!75!black,
    fonttitle=\bfseries,
    breakable,
    enhanced,
    fontlower=\small\ttfamily,
    before upper=\small\ttfamily
]
\textcolor{blue}{\{Context includes: $W_{Local}$, global frontier trajectories $\mathcal{F}$, previous exploration trajectories, and the current trajectory.\}}\\
==================================================\\
\\
Step 1:\\
Observation: \textcolor{blue}{\{current observation\}}\\
Score: \textcolor{blue}{\{current score\}}\\
Valid actions: \textcolor{blue}{\{comma-separated list of valid actions\}}\\
\\
What is your next move?
\end{tcolorbox}

\begin{tcolorbox}[
    title={User Prompt (subsequent steps)},
    colback=gray!5!white,
    colframe=gray!75!black,
    fonttitle=\bfseries,
    breakable,
    enhanced,
    fontlower=\small\ttfamily,
    before upper=\small\ttfamily
]
Step \textcolor{blue}{\{N\}}:\\
Observation: \textcolor{blue}{\{current observation\}}\\
Score: \textcolor{blue}{\{current score\}}\\
Valid actions: \textcolor{blue}{\{comma-separated list of valid actions\}}\\
\\
What is your next move?
\end{tcolorbox}

%% file: sections/examples/gwm_example.tex
\subsection{Global World Model}
\label{appendix:example_gwm_zork1}

\begin{tcolorbox}[
    title={Global World Model Example (Zork1)},
    colback=orange!5!white,
    colframe=orange!75!black,
    fonttitle=\bfseries,
    breakable,
    enhanced,
    fontlower=\small\ttfamily,
    before upper=\small\ttfamily
]
\textbf{Strategic Analysis of Game Trajectories}\\
\\
==================================================\\
\\
\textbf{1. FRONTIER \& EXPLORATION STATUS}\\
\\
\textbf{Successfully Reached Areas:}\\
- Starting point: Mailbox and House exterior (north, south, east, west of house)\\
- Forest Path and Clearing with grating\\
- Up a Tree (bird's nest with egg and leaflet)\\
- Behind House (window entry)\\
- Kitchen and Living Room inside the house\\
- Attic (Trajectory 1)\\
- Cellar and Troll Room (Trajectory 1 \& 2)\\
\\
\textbf{Unexplored or Inaccessible Areas:}\\
- Upstairs beyond the dark staircase (pitch black, dangerous)\\
- West door with gothic lettering (nailed shut)\\
- Crawlway south in cellar and forbidding hole west in Troll Room\\
- Steep metal ramp in cellar (unclimbable)\\
- Chasm area (Trajectory 1 ends in death)\\
- Any further passages beyond troll or trap door\\
\\
==================================================\\
\\
\textbf{2. GAME CHECKPOINTS \& PROGRESS}\\
\\
\textbf{Key Milestones:}\\
- Obtaining jewel-encrusted egg and leaflet from mailbox/tree\\
- Using egg and leaflet in grating to unlock cellar access\\
- Entering house via window and acquiring items (food, garlic, water, lantern, sword, rope, knife)\\
- Opening trap door and descending into cellar\\
- Defeating or engaging troll in cellar\\
\\
\textbf{Items/Abilities Unlocking New Areas:}\\
- Egg + Leaflet → Grating → Cellar entrance\\
- Lantern (turned on) → Safe navigation in dark cellar\\
- Sword (elvish, glowing) → Combat with troll\\
- Rope and knife (potential utility, not fully leveraged yet)\\
- Food, garlic, water (likely survival or puzzle items)\\
\\
\textbf{Current Phase:} Early-to-mid game - exploration of house and cellar, initial combat encounters\\
\\
==================================================\\
\\
\textbf{3. BOTTLENECKS \& CHALLENGES}\\
\\
\textbf{Common Stuck Points:}\\
- Combat with troll (Trajectories 1 \& 2 end in death or near death)\\
- Navigating dark areas without proper light (pitch black upstairs)\\
- Attempting risky actions (jumping over chasm leads to death)\\
- Locked or barred doors (gothic door, trap door closing behind player)\\
\\
\textbf{Obstacles Blocking Progress:}\\
- Physical barriers (nailed door, barred trap door)\\
- Hostile NPC (troll)\\
- Environmental hazards (darkness, chasm)\\
- Limited inventory or missing key items\\
\\
\textbf{Missing Resources/Knowledge:}\\
- Effective combat strategy or stronger weaponry to defeat troll safely\\
- Means to reopen or bypass barred trap door\\
- Safe traversal methods for chasm or dark upstairs\\
- Possible puzzle solutions involving rope, knife, or other items\\
\\
==================================================\\
\\
\textbf{4. REWARD STRUCTURE}\\
\\
\textbf{When Points Are Earned:}\\
- +5 for taking the egg\\
- +10 for acquiring food and water items in kitchen\\
- +25 for entering cellar (significant milestone)\\
- +5 for moving north from troll room to passage\\
- -10 on death and respawn (penalty)\\
\\
\textbf{Highest Reward Actions:}\\
- Descending into cellar (+25)\\
- Collecting key items early (+5 to +10)\\
- Progressing past major checkpoints\\
\\
\textbf{Scoring Patterns:}\\
- Early exploration yields small incremental rewards\\
- Major area transitions yield large rewards\\
- Death penalizes score, encouraging cautious play\\
\\
==================================================\\
\\
\textbf{5. NEXT INVESTIGATION GOALS}\\
\\
\textbf{Specific Objectives:}\\
- Develop reliable strategy to defeat or bypass troll without dying\\
- Investigate use of rope and knife for puzzles or combat\\
- Explore crawlway south and forbidding hole west in troll room\\
- Find safe method to cross or bypass chasm\\
- Attempt to open or circumvent nailed gothic door\\
\\
\textbf{Most Promising Unexplored Areas:}\\
- Crawlway south and forbidding hole west in cellar/troll room\\
- Upstairs dark staircase (with lantern or other light source)\\
- Beyond barred trap door if it can be reopened\\
- Areas beyond chasm once safe crossing method found
\end{tcolorbox}

%% file: sections/examples/lwm_example.tex
\subsection{Local World Model}
\label{appendix:example_lwm_zork1}

\begin{tcolorbox}[
    title={Local World Model Example (Zork1)},
    colback=blue!5!white,
    colframe=blue!75!black,
    fonttitle=\bfseries,
    breakable,
    enhanced,
    fontlower=\small\ttfamily,
    before upper=\small\ttfamily
]
Based on detailed exploration attempts, here are KEY STATES with discovered advantages:\\
\\
==================================================\\
\\
\textbf{STATE: The Troll Room}\\
\textit{Small room with passages east/south, hole west; troll blocks all exits; sword glows}\\
\\
\textbf{ADVANTAGES:}\\
- "hit troll with sword" → kills troll, clears all exits for exploration (score: 40)\\
- "take axe" after troll death → obtains valuable weapon for later puzzles\\
- avoid engaging without glowing sword → likely failure\\
- optimal: kill troll first, then take axe (mandatory to proceed)\\
\\
==================================================\\
\\
\textbf{STATE: Maintenance Room (Flood Control Dam \#3)}\\
\textit{Room with colored buttons, tool chests, wrench, screwdriver, toothpaste, leaking pipe}\\
\\
\textbf{ADVANTAGES:}\\
- "take wrench, screwdriver, tube" → essential tools for environment interaction\\
- "push blue button" → triggers leak, raises water level to access new areas\\
- "push red button" → toggles lights, affects water level\\
- avoid throwing lantern → breaks critical light source\\
- optimal: collect tools → manage buttons → control water without drowning\\
\\
==================================================\\
\\
\textbf{STATE: Temple / Torch Room / Dome Room / Altar}\\
\textit{Large temple with inscriptions; dome with railing; rope for descent; ivory torch; brass bell; gold coffin}\\
\\
\textbf{ADVANTAGES:}\\
- "take ivory torch" → stable light for deeper cave exploration\\
- "take bell" → key item for spirit/wraith interaction\\
- "ring bell at Entrance to Hades" → paralyzes wraiths, enables passage\\
- "blow out candles" → enables safe descent or passage\\
- optimal: acquire torch → bell → sceptre → manipulate altar → control spirits\\
\\
==================================================\\
\\
\textbf{STATE: East-West Passage / Chasm Area}\\
\textit{Narrow passage with stairs; chasm with paths; multiple routes (north/east/west/up/down)}\\
\\
\textbf{ADVANTAGES:}\\
- "east" then "north" → leads to Reservoir South and further areas\\
- "tie rope to railing" → enables safe descent into lower levels\\
- avoid getting stuck in loops → wastes moves\\
- optimal: explore chasm edges → use rope for vertical → access Dome/Torch\\
\\
==================================================\\
\\
\textbf{Cross-Cutting Insights:}\\
- Inventory Management: Strategic dropping/picking essential for critical artifacts\\
- Light Preservation: Maintaining lantern/torch crucial for dark exploration\\
- Combat Readiness: Glowing sword indicates combat opportunity (essential for progress)\\
\end{tcolorbox}